\documentclass[journal]{IEEEtran}
\ifCLASSINFOpdf
  \usepackage[pdftex]{graphicx}
  \graphicspath{{Images/}}
  \DeclareGraphicsExtensions{.pdf,.jpeg,.png}
\else
\fi

\usepackage{verbatim}
    
\usepackage{hyperref}
\usepackage[cmex10]{amsmath}
\usepackage{color}

\usepackage[normalem]{ulem}

\usepackage{array}
\usepackage{multirow}

\ifCLASSOPTIONcompsoc
 \usepackage[caption=false,font=normalsize,labelfont=sf,textfont=sf]{subfig}
\else
 \usepackage[caption=false,font=footnotesize]{subfig}
\fi
\usepackage{stfloats}

\usepackage{url}
\begin{document}
\title{RSVQA: Visual Question Answering for Remote Sensing Data}
%
%
%
\author{Sylvain Lobry,~\IEEEmembership{Member,~IEEE,}
        Diego Marcos,
        Jesse Murray,
        Devis Tuia,~\IEEEmembership{Senior Member,~IEEE}
\thanks{Sylvain Lobry, Diego Marcos, Jesse Murray and Devis Tuia are with Laboratory of Geo-Information Science and Remote Sensing, Wageningen University, The Netherlands email: work@sylvainlobry.com}
}

\markboth{Pre-print. Final version in IEEE TRANSACTIONS ON GEOSCIENCE AND REMOTE SENSING}%
{Lobry \MakeLowercase{\textit{et al.}}: RSVQA: Visual Question Answering for Remote Sensing Data}

\maketitle
\begin{small}
{ This is the pre-acceptance version, to read the final version published in the journal IEEE Transactions on Geoscience and Remote Sensing, please go to: \url{https://doi.org/10.1109/TGRS.2020.2988782}.\vspace{1em}}
\end{small}

\begin{abstract}
 This paper introduces the task of visual question answering for remote sensing data (RSVQA). Remote sensing images contain a wealth of information which can be useful for a wide range of tasks including land cover classification, object counting or detection. However, most of the available methodologies are task-specific, thus inhibiting generic and easy access to the information contained in remote sensing data. As a consequence, accurate remote sensing product generation still requires expert knowledge. With RSVQA, we propose a system to extract information from remote sensing data that is accessible to every user: we use questions formulated in natural language and use them to interact with the images. With the system, images can be queried to obtain high level information specific to the image content or relational dependencies between objects visible in the images. Using an automatic method introduced in this article, we built two datasets (using low and high resolution data) of image/question/answer triplets. The information required to build the questions and answers is queried from OpenStreetMap (OSM). The datasets can be used to train (when using supervised methods) and evaluate models to solve the RSVQA task. We report the results obtained by applying a model based on Convolutional Neural Networks (CNNs) for the visual part and on a Recurrent Neural Network (RNN) for the natural language part to this task. The model is trained on the two datasets, yielding promising results in both cases.
\end{abstract}


%
\IEEEpeerreviewmaketitle

\section{Introduction}
\label{sec:intro}
\IEEEPARstart{R}{emote} sensing data is widely used as an \emph{indirect} source of information. From land cover/land use to crowd estimation, environmental or urban area monitoring, remote sensing images are used in a wide range of tasks of high societal relevance. For instance, remote sensing data can be used as a source of information for 6 of the 17 sustainable development goals as defined by the United Nations \cite{anderson2017earth}.  Due to the critical nature of the problems that can be addressed using remote sensing data, significant  effort  has been made to increase its availability in the last decade. For instance, Sentinel-2 satellites provide multispectral data with a relatively short revisiting time, in open-access. However, while substantial effort has been dedicated to improving the means of direct information extraction from Sentinel-2 data in the framework of a given task (e.g. classification \cite{8003345, 8697135}), the ability to use remote sensing data as a \emph{direct} source of information is currently limited to experts within the remote sensing and computer vision communities. This constraint, imposed by the technical nature of the task, reduces both the scale and variety of the problems that could be addressed with such information as well as the number of potential end-users. This is particularly true when targeting specific applications (detecting particular objects, \emph{e.g.} thatched roofs or buildings in a developing country  \cite{vargas2019correcting}) which would today call for important research efforts. The targeted tasks are often multiple and changing in the scope of a project calls for strong expert knowledge, limiting the information which can be extracted from remote sensing data. To address these constraints, we introduce the problem of visual question answering (VQA) for remote sensing data.

\begin{figure}
    \centering
    \includegraphics[width=\columnwidth]{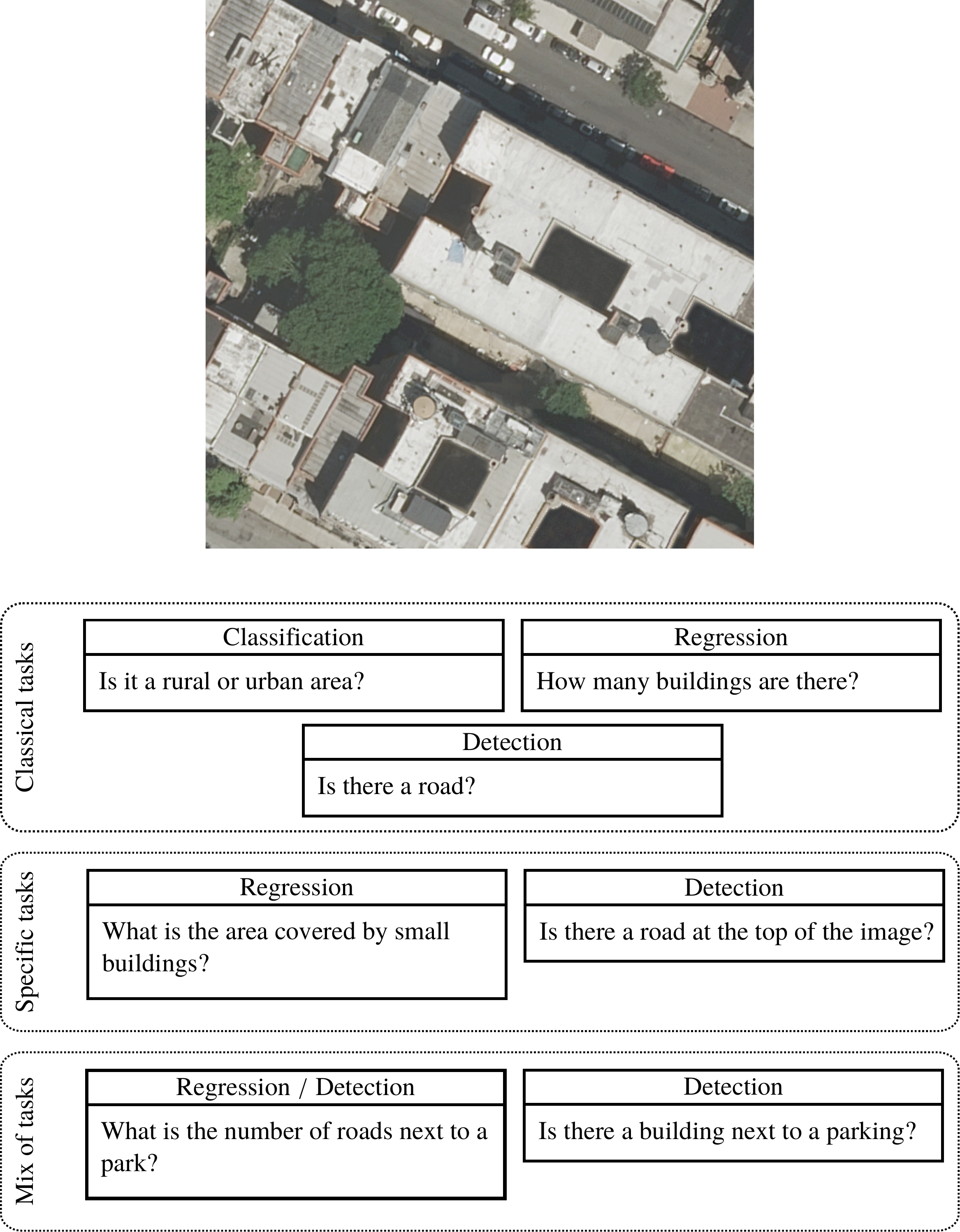}
    \caption{Example of tasks achievable by a visual question answering model for remote sensing data.}
    \label{fig:RSVQA_Sample_questions}
\end{figure}

VQA is a new task in computer vision, introduced in its current form by \cite{VQA}. The objective of VQA is to answer a free-form and open-ended question about a given image. As the questions can be unconstrained, a VQA model applied to remote sensing data could serve as a generic solution to classical problems involving remote sensing data (e.g. "Is there a thatched roof in this image?" for thatched roof detection), but also very specific tasks involving relations between objects of different nature (e.g. "Is there a thatched roof on the right of the river?"). Examples of potential questions are shown in \autoref{fig:RSVQA_Sample_questions}.\\

To the best of our knowledge, this is the first time (after the first exploration in \cite{VQAIGARSS}) that VQA has been applied to extract information from remote sensing data. It builds on the task of generating descriptions of images through combining image and natural language processing to provide the user with easily accessible, high-level semantic information.  These descriptions are then used for image retrieval and intelligence generation \cite{7891049}. As seen in this introduction, VQA systems rely on the recent advances in deep learning. Deep learning based methods, thanks to their ability to extract high-level features, have been successfully developed for remote sensing data as reviewed in \cite{8113128}. Nowadays, this family of methods is used to tackle a variety of tasks; for scene classification, an early work by \cite{hu2015transferring} evaluated the possibility to adapt networks pre-trained on large natural image databases (such as ImageNet \cite{imagenet_cvpr09}) to classify hyperspectral remote sensing images. More recently, \cite{8454883} used an intermediate high level representation using recurrent attention maps to classify images. Object detection is also often approached using deep learning methods. To this effect, \cite{xia2018dota} introduced an object detection dataset and evaluated classical deep learning approaches. Methods taking into account the specificity of remote sensing data have been developed, such as \cite{8651485} which proposed to modify the classical approach by generating rotatable region proposal which are particularly relevant for top-view imagery. Deep learning methods have also been developed for semantic segmentation. In \cite{volpi2016dense}, the authors evaluated different strategies for segmenting remote sensing data. More recently, a contest organized on the dataset of building segmentation created by \cite{maggiori2017can} has motivated the development of a number of new methods to improve results on this task \cite{huang2018large}. Similarly, \cite{demir2018deepglobe} introduced a contest including three tasks: road extraction, building detection and land cover classification. Best results for each challenge were obtained using deep neural networks: \cite{zhou2018d}, \cite{hamaguchi2018building}, \cite{tian2018dense}.

Natural language processing has also been used in remote sensing. For instance, \cite{8128075} used a convolutional neural network (CNN) to generate classification probabilities for a given image, and used a recurrent neural network (RNN) to generate its description. In a similar fashion,  \cite{7891049} used a CNN to obtain a multi semantic level representation of an image (object, land class, landscape) and generate a description using a simple static model. More recently, \cite{zhang2019description} uses an encoder/decoder type of architecture where a CNN encodes the image and a RNN decodes it to a textual representation, while \cite{wang2019semantic} projects the textual representation and the image to a common space. While these works are use cases of natural language processing, they do not enable interactions with the user as we propose with VQA.\\

A VQA model is generally made of 4 distinct components: 1) a visual feature extractor, 2) a language feature extractor, 3) a fusion step between the two modalities and 4) a prediction component.
Since VQA is a relatively new task, an important number of methodological developments have been published in both the computer vision and natural language processing communities during the past 5 years, reviewed in \cite{DBLP:journals/corr/WuTWSDH16}. VQA models are able to benefit from advances in the computer vision and automatic language processing communities for the features extraction components. However, the multi-modal fusion has been less explored and therefore, an important amount of work has been dedicated to this step. First VQA models relied on a non-spatial fusion method, \emph{i.e.} a point-wise multiplication between the visual and language feature vectors \cite{VQA}. Being straightforward, this method does not allow every component  from both feature vectors to interact with each other. This interaction would ideally be achieved by multiplying the first feature vector by the transpose of the other, but this operation would be computationally intractable in practice. Instead, \cite{fukui2016multimodal} proposed a fusion method which first selects relevant visual features based on the textual feature (attention step) and then, combines them with the textual feature. In \cite{ben2017mutan}, the authors used Tucker decomposition to achieve a similar purpose in one step. While these attention mechanisms are interesting for finding visual elements aligned with the words within the question, they require the image to be divided in a regular grid for the computation of the attention, and this is not suitable to objects of varying size. A solution is presented in \cite{anderson2018bottom}, which learns an object detector to select relevant parts of the image. In this research, we use a non-spatial fusion step to keep the model part relatively simple. Most traditional VQA works are designed for a specific dataset, either composed of natural images (with questions covering an unconstrained range of topics) or synthetic images. While interesting for the methodological developments that they have facilitated, these datasets limit the potential applications of such systems to other problems. Indeed, it has been shown in \cite{DBLP:journals/corr/abs-1903-00366} that VQA models trained on a specific dataset do not generalize well to other datasets. This \emph{generalization gap} raises questions concerning the applicability of such models to specific tasks.

A notable use-case of VQA is helping visually impaired people through natural language interactions \cite{bigham2010vizwiz}. Images acquired by visually impaired people represent an important domain shift, and as such a challenge for the applicability of VQA models. In \cite{DBLP:journals/corr/abs-1802-08218}, the authors confirm that networks trained on generic datasets do not generalize to their specific one. However, they manage to obtain much better results by fine-tuning or training models from scratch on their task-specific dataset.\\

In this study, we propose a new application for VQA, specifically for the interaction with remote sensing images. To this effect, we propose the first remote sensing-oriented VQA datasets, and evaluate the applicability of this task on remote sensing images. We propose a method to automatically generate remote sensing-oriented VQA datasets from already available human annotations in \autoref{sec:dataset} and generate two datasets. We then use this newly-generated data to train our proposed RSVQA model with a non-spatial fusion step described in \autoref{sec:model}. Finally, the results are evaluated and discussed in \autoref{sec:results}.

Our contribution are the following:
\begin{itemize}
    \item a method to generate remote sensing-oriented VQA datasets;
    \item 2 datasets;
    \item the proposed RSVQA model.
\end{itemize}
This work extends the preliminary study of \cite{VQAIGARSS} by considering and disclosing a second larger dataset consisting of very high resolution images. This second dataset helps testing the spatial generalization capability of VQA and provides an extensive discussion highlighting remaining challenges.
The method to generate the dataset, the RSVQA model and the two datasets are available on \url{https://rsvqa.sylvainlobry.com/}. 

\section{Datasets}
\label{sec:dataset}

\subsection{Method}
\label{ssec:dataset_method}
\begin{figure}
    \centering
    \subfloat[Question construction procedure. Dash lines represent optional paths]{\includegraphics[width=\columnwidth]{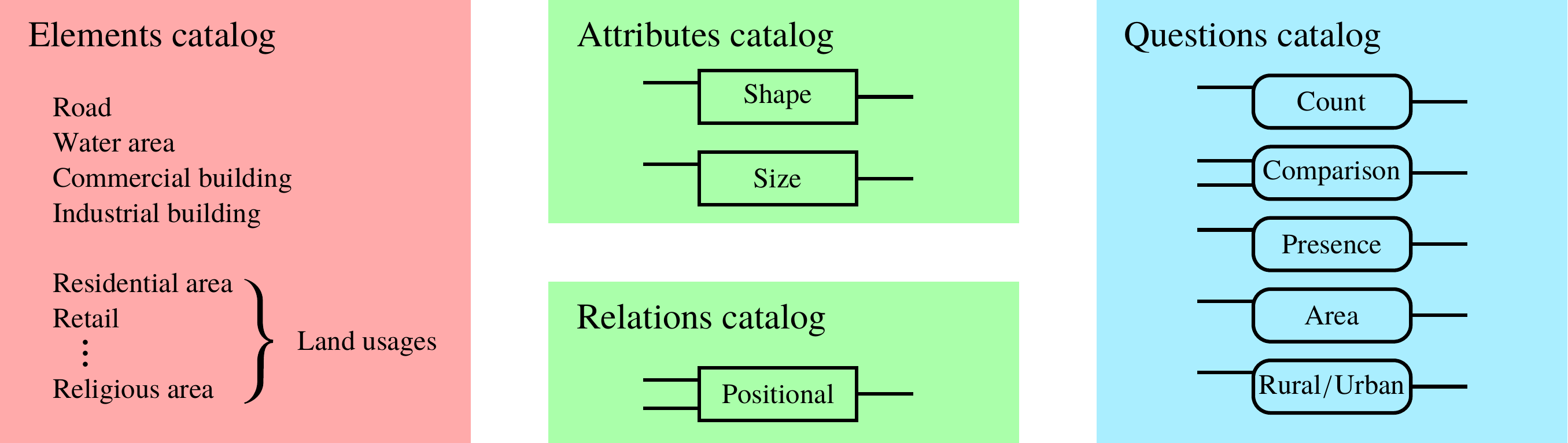}}\vspace{1em}\\
    \subfloat[Construction path for sample questions.]{\includegraphics[width=\columnwidth]{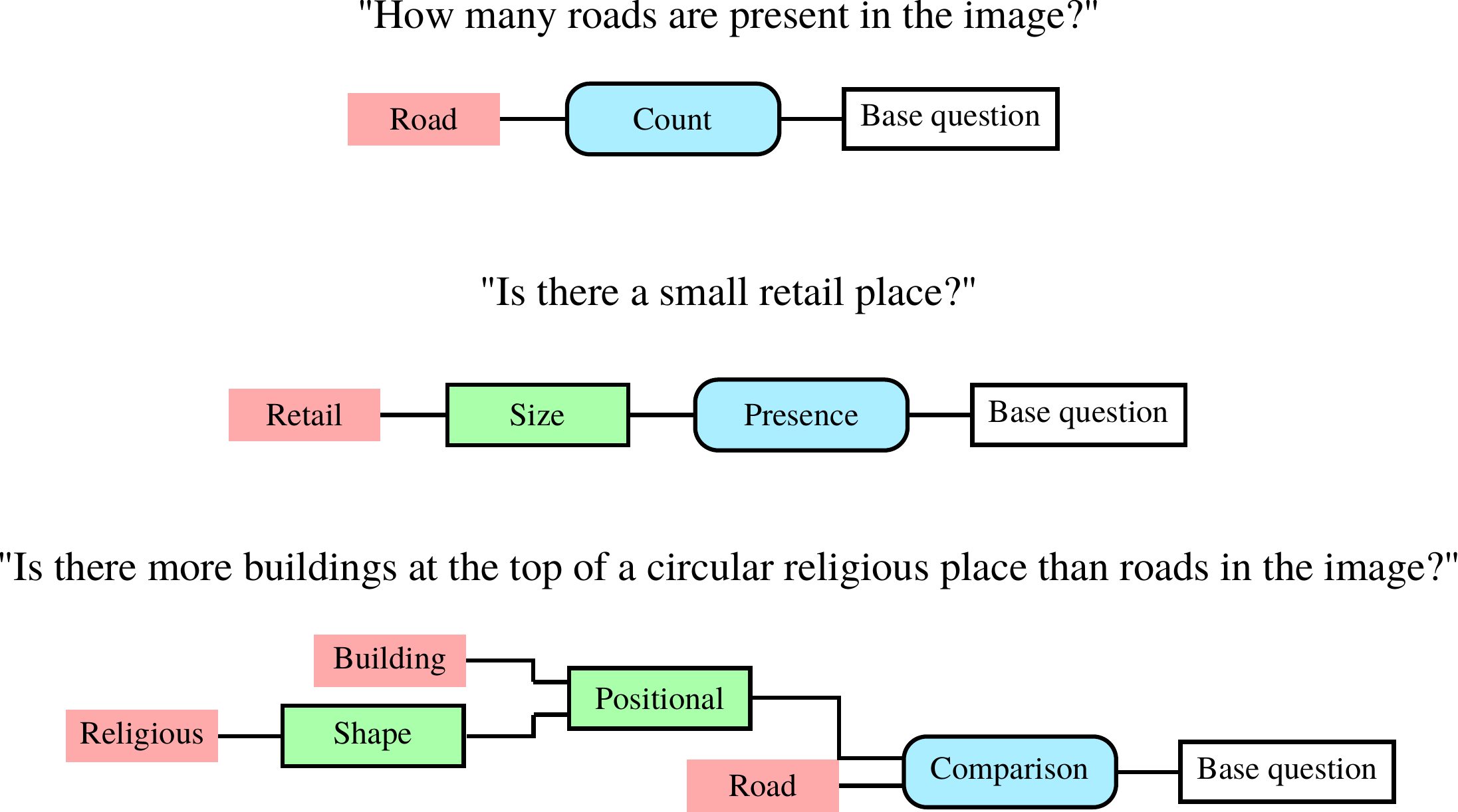}}\\
    \caption{Illustration of the question construction procedure.}
    \label{fig:QuestionConstruction}
\end{figure}
As seen in the introduction, a main limiting factor for VQA is the availability of task-specific datasets. As such, we aim at providing a collection of remote sensing images with questions and answers associated to them. To do so, we took inspiration from \cite{DBLP:journals/corr/JohnsonHMFZG16}, in which the authors build a dataset of question/answer pairs about synthetic images following an automated procedure. However, in this study we are interested in real data (discussed in \autoref{ssec:dataset_data}). Therefore, we use the openly accessible OpenStreetMap data containing geo-localized information provided by volunteers. By leveraging this data, we can automatically extract the information required to obtain question/answer pairs relevant to real remotely sensed data and create a dataset made of (image, question, answer) triplets.\\

The first step of the database construction is to create the questions. The second step is to compute the answers to the questions, using the OSM features belonging to the image footprint. Note that multiple question/answer pairs are extracted for each image.

~\\
\subsubsection{Question contruction}
Our method to construct the questions is illustrated in \autoref{fig:QuestionConstruction}. It consists of four main components:
\begin{enumerate}
    \item choice of an element category (highlighted in red in \autoref{fig:QuestionConstruction}(a));
    \item application of attributes to the element (highlighted in green in \autoref{fig:QuestionConstruction}(a));
    \item selection based on the relative location to another element (highlighted in green in \autoref{fig:QuestionConstruction}(a))
    \item construction of the question (highlighted in blue in \autoref{fig:QuestionConstruction}(a)).
\end{enumerate}
Examples of question constructions are shown in \autoref{fig:QuestionConstruction}(b). These four components are detailed in the following.\\

\noindent
\textbf{Element category selection}:
First, an element category is randomly selected from the element catalog. This catalog is built by extracting the elements from one of the following OSM layers: \emph{road}, \emph{water area}, \emph{building} and \emph{land use}. While roads and water areas are directly treated as elements, buildings and land use related objects are defined based on their "type" field, as defined in the OSM data specification. Examples of land use objects include residential area, construction area, religious places, \ldots Buildings are divided in two categories: commercial (e.g. retail, supermarket, ...) and residential (e.g. house, apartments, \ldots).

~\\
\noindent
\textbf{Attributes application}:
The second (optional) step is to refine the previously selected element category. To do so, we randomly select from one of the two possible attribute categories:
\begin{itemize}
    \item \textbf{Shape}:
    each element can be either square, rectangular or circular. Whether an element belongs to one of these shape types is decided based on basic geometrical properties (i.e. hard thresholds on area-to-perimeter ratio and area-to-circumscribed circle area ratio).
    \item \textbf{Size}:
    using hard thresholds on the surface area, elements can be considered "small", "medium" or "large". As we are interested in information at different scales in the two datasets, we use different threshold values, which are described in \autoref{tab:threshsizes}.
\end{itemize}
\begin{table}
    \centering
    \begin{tabular}{|c|c|c|c|}
        \hline
        Scale & Small & Medium & Large \\
        \hline
        Low resolution & $<$ 3000m$^2$ & $<$ 10000m$^2$ & $\geq$ 10000m$^2$\\
        \hline
        High resolution & $<$ 100m$^2$ & $<$ 500m$^2$ & $\geq$ 500m$^2$\\
        \hline
    \end{tabular}
    \caption{Thresholds for size attributes according to the dataset scale. When dealing with low resolution data, visible objects of interest are larger. To deal with this disparity, we adapt the size thresholds to the resolution of the images.}
    \label{tab:threshsizes}
\end{table}

~\\
\noindent
\textbf{Relative position}:
Another possibility to refine the element is to look at its relative position compared to another element. We define 5 relations: "left of", "top of", "right of", "bottom of", "next to". Note that these relative positions are understood in the image space (i.e. geographically). The special case of "next to" is handled as a hard threshold on the relative distance between the two objects (less than 1000m). When looking at relative positions, we select the second element following the procedure previously defined.

~\\
\noindent
\textbf{Question construction}:
At this point of the procedure, we have an element (e.g. road), with an optional attribute (e.g. small road) and an optional relative position (e.g. small road on the left of a water area). The final step is to generate a "base question" about this element. We define 5 types of questions of interest ("Question catalog" in \autoref{fig:QuestionConstruction}(a)), from which a specific type is randomly selected to obtain a base question. For instance, in the case of comparison questions, we randomly choose among  "less than", "equals to" and "more than" and construct a second element.

This base question is then turned into a natural language question using pre-defined templates for each question type and object. For some question types (e.g. count), more than one template is defined (e.g. 'How many \underline{\hspace{0.5cm}} are there?', 'What is the number of \underline{\hspace{0.5cm}}?' or 'What is the amount of \underline{\hspace{0.5cm}}?'). In this case, the template to be used is randomly selected. The stochastic process ensures the diversity, both in the question types and the question templates used.

~\\
\subsubsection{Answer construction}:
To obtain the answer to the constructed question, we extract the objects from the OSM database corresponding to the image footprint. The objects $b$ corresponding to the element category and its attributes are then selected and used depending on the question type: 
\begin{itemize}
    \item \textbf{Count}:
    In the case of counting, the answer is simply the number of objects $b$.
    \item \textbf{Presence}:
    A presence question is answered by comparing the number of objects $b$ to 0.
    \item \textbf{Area}:
    The answer to a question about the area is the sum of the areas of the objects $b$.
    \item \textbf{Comparison}:
    Comparison is a specific case for which a second element and the relative position statement is needed. This question is then answered by comparing the number of objects $b$ to the ones of the second element.
    \item \textbf{Rural/Urban}:
    The case of rural/urban questions is handled in a specific way. In this case, we do not create a specific element, but rather count the number of buildings (both commercial or residential). This number of buildings is then thresholded to a predefined number depending on the resolution of the input data (to obtain a density) to answer the question. Note that we are using a generic definition of rural and urban areas but this can be easily adapted using the precise definition of each country.
\end{itemize}

\subsection{Data}
\label{ssec:dataset_data}
Following the method presented in \autoref{ssec:dataset_method}, we construct two datasets with different characteristics.
~\vspace{1em}\\
\textbf{Low resolution (LR)}:
\begin{figure}
    \centering
    \includegraphics[width=.9\columnwidth]{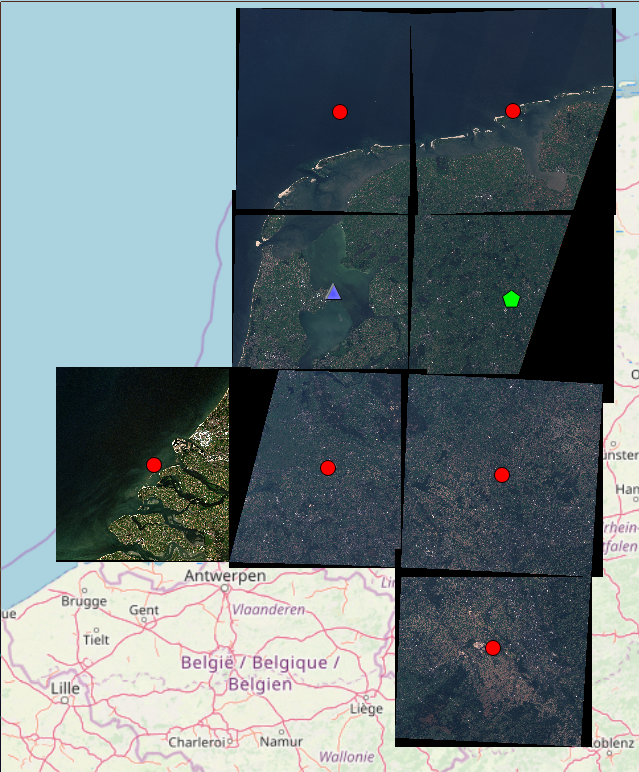}
    \caption{Images selected for the LR dataset over the Netherlands. Each point represent one Sentinel-2 image which was later split into tiles. Red points represent training samples, green pentagon represents the validation image, and blue triangle is for the test image. Note that one training image is not visible (as it overlaps with the left-most image).}
    \label{fig:ExtentLR}
\end{figure}
this dataset is based on Sentinel-2 images acquired over the Netherlands. Sentinel-2 satellites provide 10m resolution (for the visible bands used in this dataset) images with frequent updates (around 5 days) at a global scale. These images are openly available through ESA's Copernicus Open Access Hub\footnote{\url{https://scihub.copernicus.eu/}}.

To generate the dataset, we selected 9 Sentinel-2 tiles covering the Netherlands with a low cloud cover (selected tiles are shown in \autoref{fig:ExtentLR}). These tiles were divided in $772$ images of size $256\times 256$ (covering $6.55km^2$) retaining the RGB bands. From these, we constructed $77'232$ questions and answers following the methodology presented in \autoref{ssec:dataset_method}. We split the data in a training set ($77.8\%$ of the original tiles), a validation set ($11.1\%$) and a test set ($11.1\%$) at the tile level (the spatial split is shown in \autoref{fig:ExtentLR}). This allows to limit spatial correlation between the different splits.

~\vspace{1em}\\
\textbf{High resolution (HR)}:
\begin{figure}
    \centering
    \includegraphics[width=.9\columnwidth]{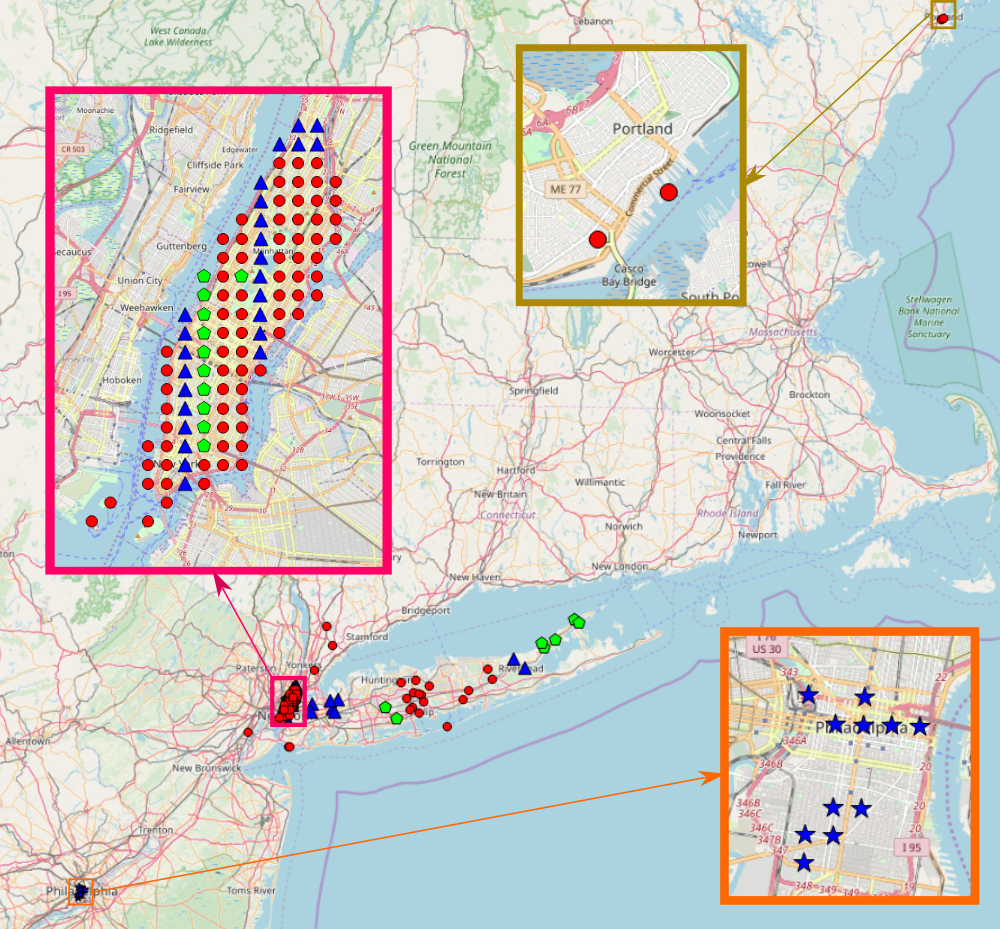}
    \caption{Extent of the HR dataset with a zoom on the Portland, Manhattan (New York City) and Philadelphia areas. Each point represent one image (generally of size $5000\times5000$) which was later split into tiles. The images cover the New York City/Long Island region, Philadelphia and Portland. Red points represent training samples, green pentagons represent validation samples, and blue indicators are for the test sets (blue triangles for test set 1, blue stars for test set 2).}
    \label{fig:ExtentUSGS}
\end{figure}
this dataset uses 15cm resolution aerial RGB images extracted from the High Resolution Orthoimagery (HRO) data collection of the USGS. This collection covers most urban areas of the USA, along with a few areas of interest (e.g. national parks). For most areas covered by the dataset, only one tile is available with acquisition dates ranging from year 2000 to 2016, with various sensors. The tiles are openly accessible through USGS' EarthExplorer tool\footnote{\url{https://earthexplorer.usgs.gov/}}.

From this collection, we extracted $161$ tiles belonging to the North-East coast of the USA (see \autoref{fig:ExtentUSGS}) that were split into $10'659$ images of size $512\times 512$ (each covering $5898m^2$). We constructed $1'066'316$ questions and answers following the methodology presented in \autoref{ssec:dataset_method}. We split the data in a training set ($61.5\%$ of the tiles), a validation set ($11.2\%$), and test sets ($20.5\%$ for test set 1, $6.8\%$ for test set 2). As it can be seen in \autoref{fig:ExtentUSGS}, test set 1 covers similar regions as the training and validation sets, while test set 2 covers the city of Philadelphia, which is not seen during the training. Note that this second test set also uses another sensor (marked as unknown on the USGS data catalog), not seen during training.

\begin{figure*}
    \centering
    \subfloat[Distribution of answers for the LR dataset]{\includegraphics[width=.88\columnwidth]{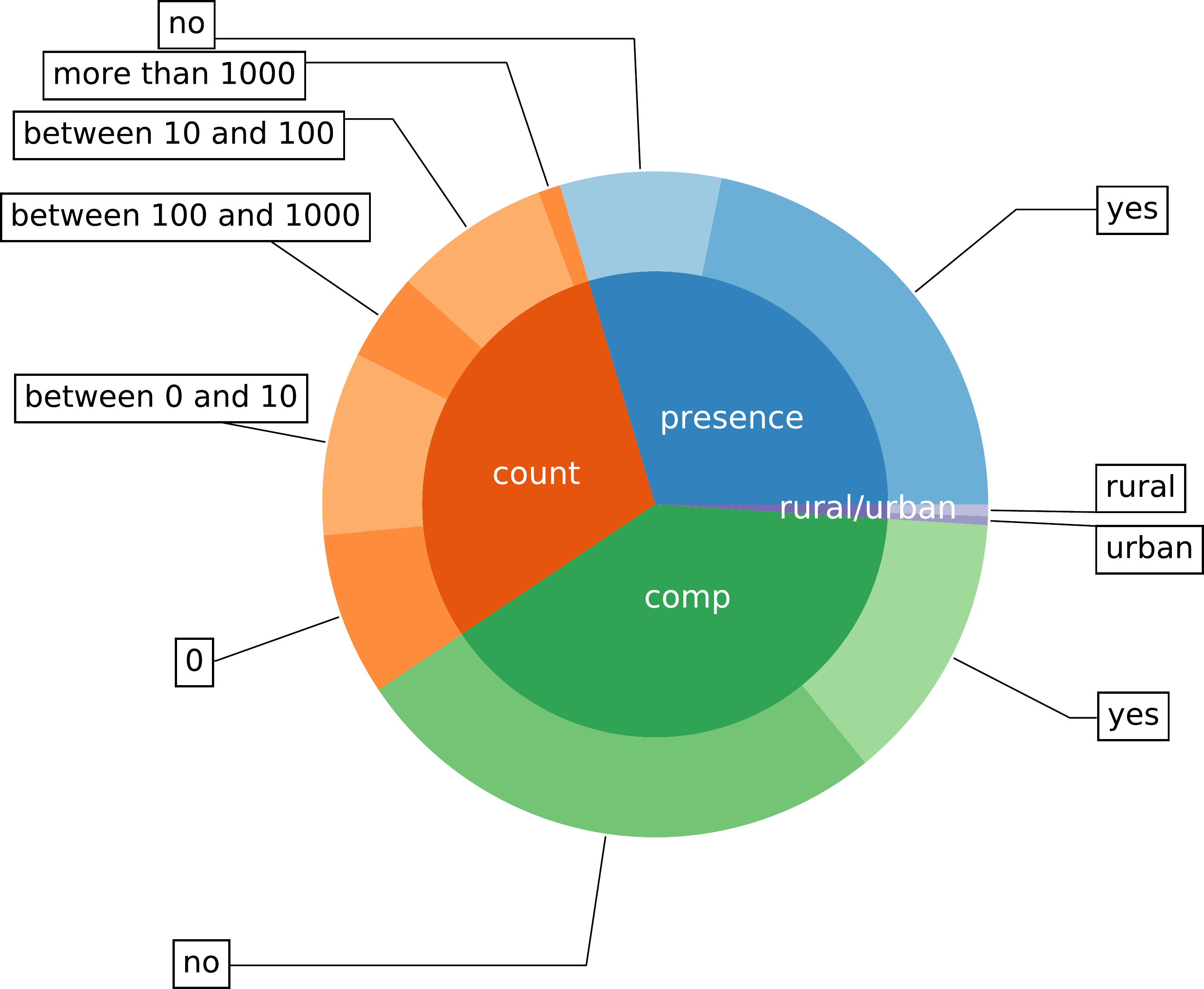}}\hspace{1em}
    \subfloat[Distribution of answers for the HR dataset (numerical answers are ordered, and 0 is the most frequent)]{\includegraphics[width=\columnwidth]{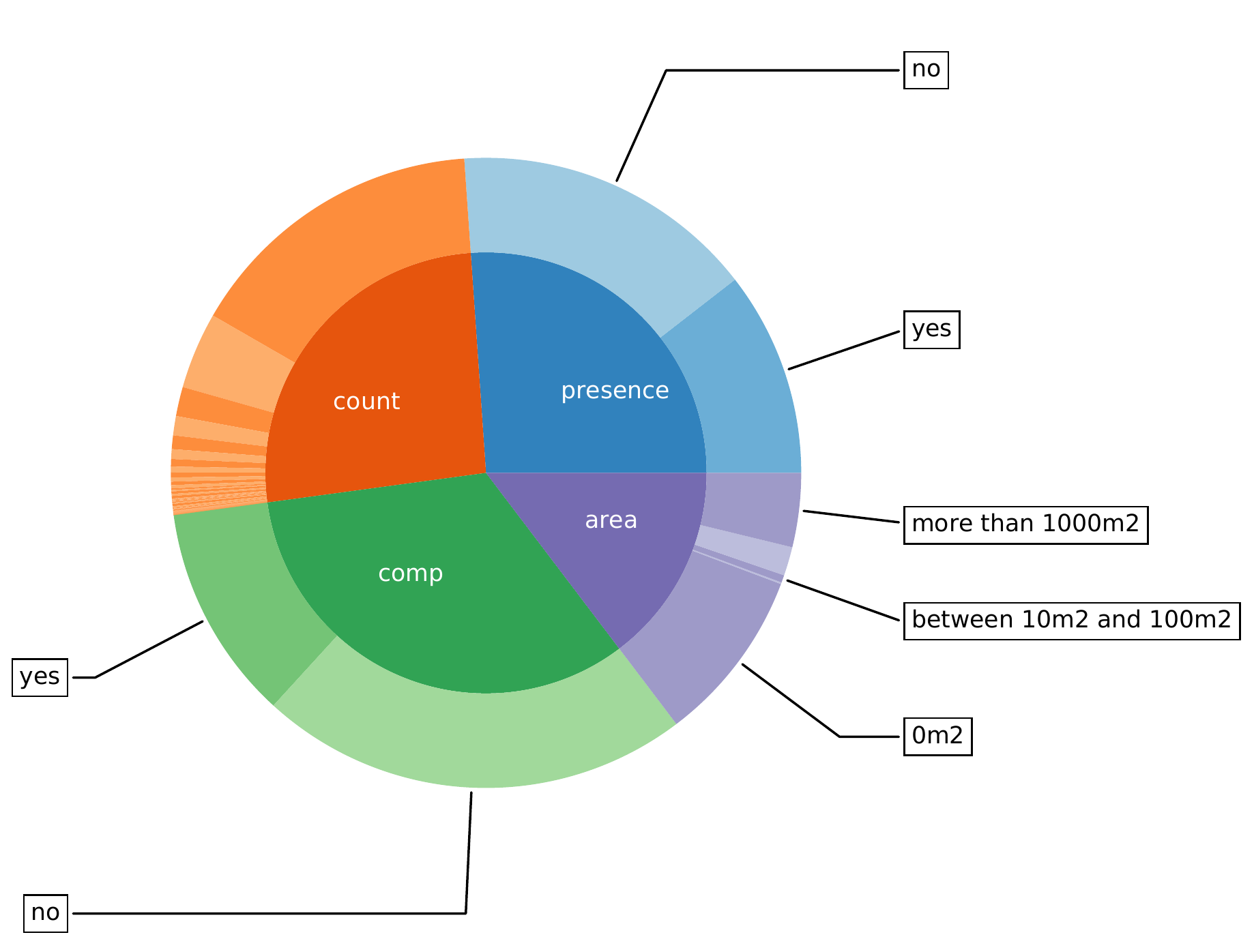}}
    \caption{\label{fig:answer_distribution}Distributions of answers in the Low resolution (LR) and High resolution (HR) datasets.}
\end{figure*}

~\\
\noindent
\textbf{Differences between the two datasets}:\\
Due to their characteristics, the two datasets represent two different possible use cases of VQA:
\begin{itemize}
    \item The LR dataset allows for large spatial and temporal coverage thanks to the frequent acquisitions made by Sentinel-2. This characteristic could be of interest for future applications of VQA such as large scale queries (e.g. rural/urban questions) or temporal (which is out of the scope of this study). However, due to the relatively low resolution (10m), some objects can not be seen on such images (such as small houses, roads, trees, \ldots). This fact severely limits the questions to which the model could give an accurate answer.
    \item Thanks to the much finer resolution of the HR dataset, a quantity of information of interest to answer typical questions is present. Therefore, in contrast to the LR dataset, questions concerning objects' coverage or counting relatively small objects can possibly be answered from such data. However, data of such resolution is generally less frequently updated and more expensive to acquire.
\end{itemize}

Based on these differences, we constructed different types of questions for the two datasets. Questions concerning the area of objects are only asked in the HR dataset. On the other hand, questions about urban/rural area classification are only asked in the LR dataset, as the level of zoom of images from the HR dataset would prevent a meaningful answer from being provided.

\begin{figure}
    \centering
    \includegraphics[width=.7\columnwidth]{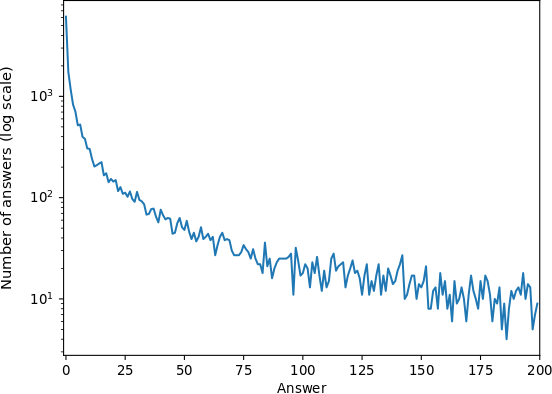}
    \caption{\label{fig:countLR}Frequencies of exact counting answers in the LR dataset. Only the left part of the histogram is shown (until 200 objects), the largest (single) count being 17139. $50\%$ of the answers are less than 7 objects in the tile.}
    \label{fig:my_label}
\end{figure}

To account for the data distributions and error margins we also quantize different answers in both datasets:
\begin{itemize}
    \item Counting in LR: as the coverage is relatively large (6.55km$^2$), the number of small objects contained in one tile can be high, giving a heavy tailed distribution for the numerical answers, as shown in \autoref{fig:countLR}. More precisely, while 26.7\% of the numerical answers are '0' and 50\% of the answers are less than '7', the highest numerical answer goes up to '17139'. In addition to making the problem complex, we can argue that allowing such a range of numerical answer does not make sense on data of this resolution. Indeed, it would be in most cases impossible to distinguish 17139 objects on an image of 65536 pixels. Therefore, numerical answers are quantized into the following categories:
    \begin{itemize}
        \item '0';
        \item 'between 1 and 10';
        \item 'between 11 and 100';
        \item 'between 101 and 1000';
        \item 'more than 1000'.
    \end{itemize}
    \item In a similar manner, we quantize questions regarding the area in the HR dataset. A great majority (60.9\%) of the answer of this type are '0m$^2$', while the distribution also presents a heavy tail. Therefore, we use the same quantization as the one proposed for counts for the LR dataset. Note that we do not quantize purely numerical answers (i.e. answers to questions of type 'count') as the maximum number of objects is 89 in our dataset. Counting answers therefore correspond to 89 classes in the model in this case (see \autoref{sec:model}).
\end{itemize}
\subsection{Discussion}
\label{ssec:discussion_data}
\noindent
\textbf{Questions/Answers distributions}:\\
We show the final distribution of answers per question type for both datasets in Figure \ref{fig:answer_distribution}. We can see that most question types (with the exception of 'rural/urban' questions in the LR dataset, asked only once per image) are close to evenly distributed by construction. The answer 'no' is dominating the answers' distribution for the HR dataset with a frequency of 37.7\%. In the LR dataset, the answer 'yes' occurs 34.9\% of the time while the 'no' frequency is 34.3\%. The strongest imbalance occurs for the answer '0' in the HR dataset (with a frequency of 60.9\% for the numerical answer). This imbalance is greatly reduced by the quantization process described in the previous paragraph.\\

\noindent
\textbf{Limitations of the proposed method}:\\
While the proposed method for image/question/answer triplets generation has the advantage of being automatic and easily scalable while using data annotated by humans, a few limitations have been observed. First, it can happen that some annotations are missing or badly registered \cite{vargas2019correcting}. Furthermore, it was not possible to match the acquisition date of the imagery to the one of OSM. The main reason being that it is impossible to know if a newly added element appeared at the same time in reality or if it was just entered for the first time in OSM. As OSM is the main source of data for our process, errors in OSM will negatively impact the accuracy of our databases. 

Furthermore, due to the templates used to automatically construct questions and provide answers, the set of questions and answers is more limited than what it is in traditional VQA datasets (9 possible answers for the LR dataset, 98 for the HR dataset).

\section{VQA Model}
\label{sec:model}
\begin{figure*}
    \centering
    \includegraphics[width=.9\textwidth]{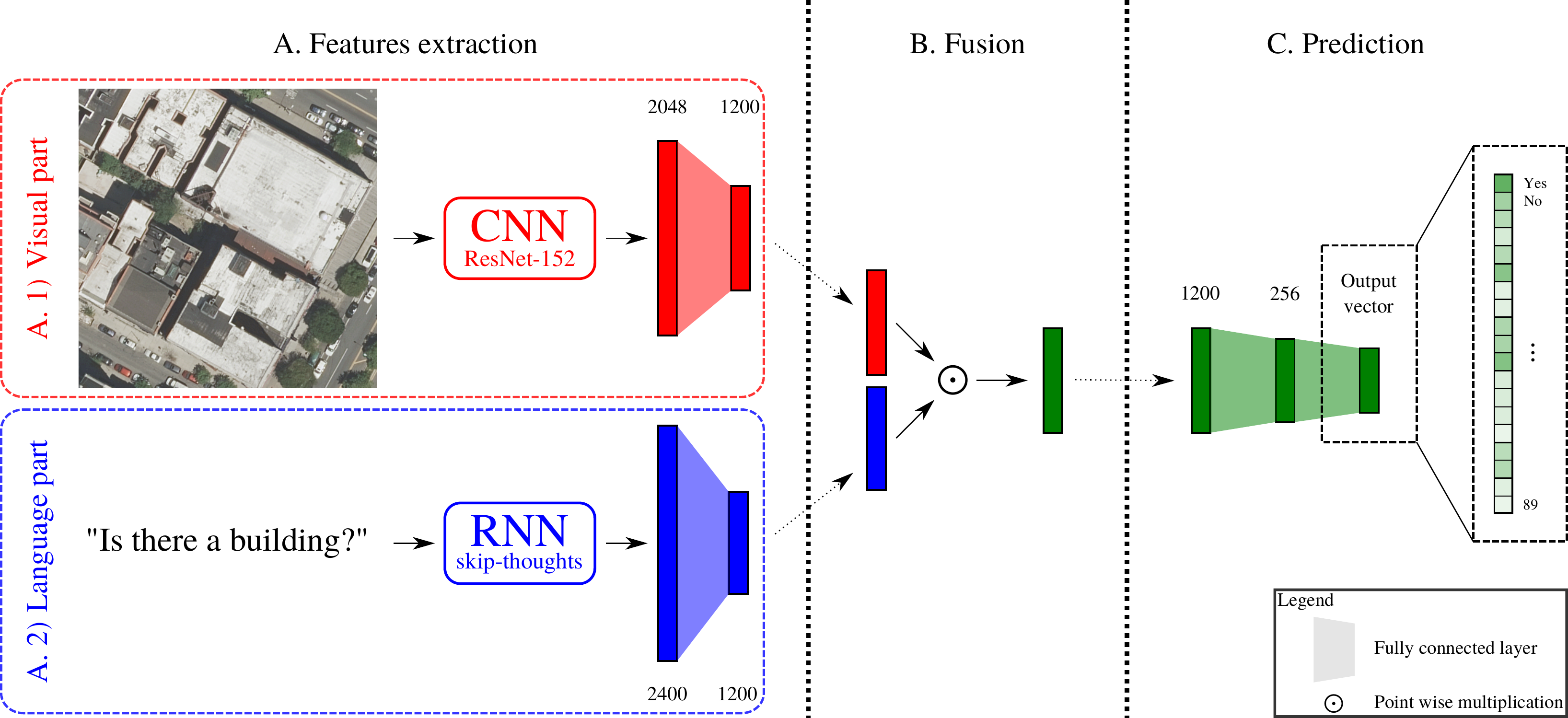}
    \caption{\label{fig:model}Framework of the proposed Visual Question Answering model.}
\end{figure*}
We investigate the difficulty of the VQA task for remote sensing using a basic VQA model based on deep learning. An illustration of the  proposed network is shown in \autoref{fig:model}. In their simple form, VQA models are composed of three parts \cite{DBLP:journals/corr/WuTWSDH16}:
\begin{itemize}
    \item[A.] feature extraction;
    \item[B.] fusion of these features to obtain a single feature vector representing both the visual information and the question;
    \item[C.] prediction based on this vector.
\end{itemize}
As the model shown in \autoref{fig:model} is learned end-to-end, the vector obtained after the fusion (in green in \autoref{fig:model}) can be seen as a joint embedding of both the image and the question which is used as an input for the prediction step. We detail each of these 3 parts in the following.

\subsection{Feature extraction}
The first component of our VQA model is the feature extraction. Its purpose is to obtain a low-dimensional representation of the information contained in the image and the question.

\subsubsection{Visual part}
To extract information from a 2D image, a common choice is to use a Convolutional Neural Network (CNN). 
Specifically, we use a Resnet-152 model \cite{DBLP:journals/corr/HeZRS15} pre-trained on ImageNet \cite{imagenet_cvpr09}. The principal motivation for this choice is that this architecture manages to avoid the undesirable \emph{degradation} problem (decreasing performance with deeper networks) by using residual mappings of the layers' inputs which are easier to learn than the common choice of direct mappings. This architecture has been succesfully used in a wide range of work in the remote sensing community (e.g. \cite{8113128, demir2018deepglobe, JURSECount}). The last average pooling layer and fully connected layer are replaced by a $1\times 1$ 2D convolution which outputs a total of 2048 features which are vectorized. A final fully connected layer is learned to obtain a 1200 dimension vector.

\subsubsection{Language part}
The feature vector is obtained using the skip-thoughts model \cite{DBLP:journals/corr/KirosZSZTUF15} trained on the BookCorpus dataset \cite{zhu2015aligning}. This model is a recurrent neural network, which aims at producing a vector representing a sequence of words (in our case, a question). To make this vector informative, the model is trained in the following way: it encodes a sentence from a book in a latent space, and tries to decode it to obtain the two adjacent sentences in the book. By doing so, it ensures that the latent space embeds semantic information. Note that this semantic information is not remote sensing specific due to the BookCorpus dataset it has been trained on. However, several works, including \cite{li2017zero}, have successfully applied non-domain specific NLP models to remote sensing. In our model, we use the encoder which is then followed by a fully-connected layer (from size 2400 elements to 1200).

\subsection{Fusion}
At this step, we have two feature vectors (one representing the image, one representing the question) of the same size. To merge them into a single vector, we use a simple strategy: a point-wise multiplication after applying the hyperbolic tangent function to the vectors' elements. While being a fixed (i.e. not learnt) operation, the end-to-end training of our model encourages both feature vectors to be comparable with respect to this operation.

\subsection{Prediction}
Finally, we project this 1200 dimensional vector to the answer space by using a MLP with one hidden layer of 256 elements. We formulate the problem as a classification task, in which each possible answer is a class. Therefore, the size of the output vector depends on the number of possible answers.

\subsection{Training procedure}
We train the model using the Adam optimizer \cite{DBLP:journals/corr/KingmaB14} with a learning rate of $10^{-5}$ until convergence (150 epochs in the case of the LR dataset, and 35 epochs in the case of the HR dataset). We use a dropout of 0.5 for every fully connected layer. Due to the difference of input size between the two datasets (HR images are 4 times larger), we use batches of 70 instances for the HR dataset and 280 for the LR dataset. Furthermore, when the questions do not contain a positional component relative to the image space (i.e. "left of", "top of", "right of" or "bottom of", see \autoref{ssec:dataset_method}), we augment the image space by randomly applying vertical and/or horizontal flipping

\section{Results and discussion}
\label{sec:results}
We report the results obtained by our model on the test sets of the LR and HR datasets. In both cases, 3 model runs have been trained and we report both the average and the standard deviation of our results to limit the variability coming from the stochastic nature of the optimization.

The numerical evaluation is achieved using the accuracy, defined in our case as the ratio of correct answers. We report the accuracy per question type (see \autoref{ssec:dataset_method}), the average of these accuracies (AA) and the overall accuracy (OA).\\

\begin{figure*}
    \centering
    \includegraphics[width=.87\textwidth]{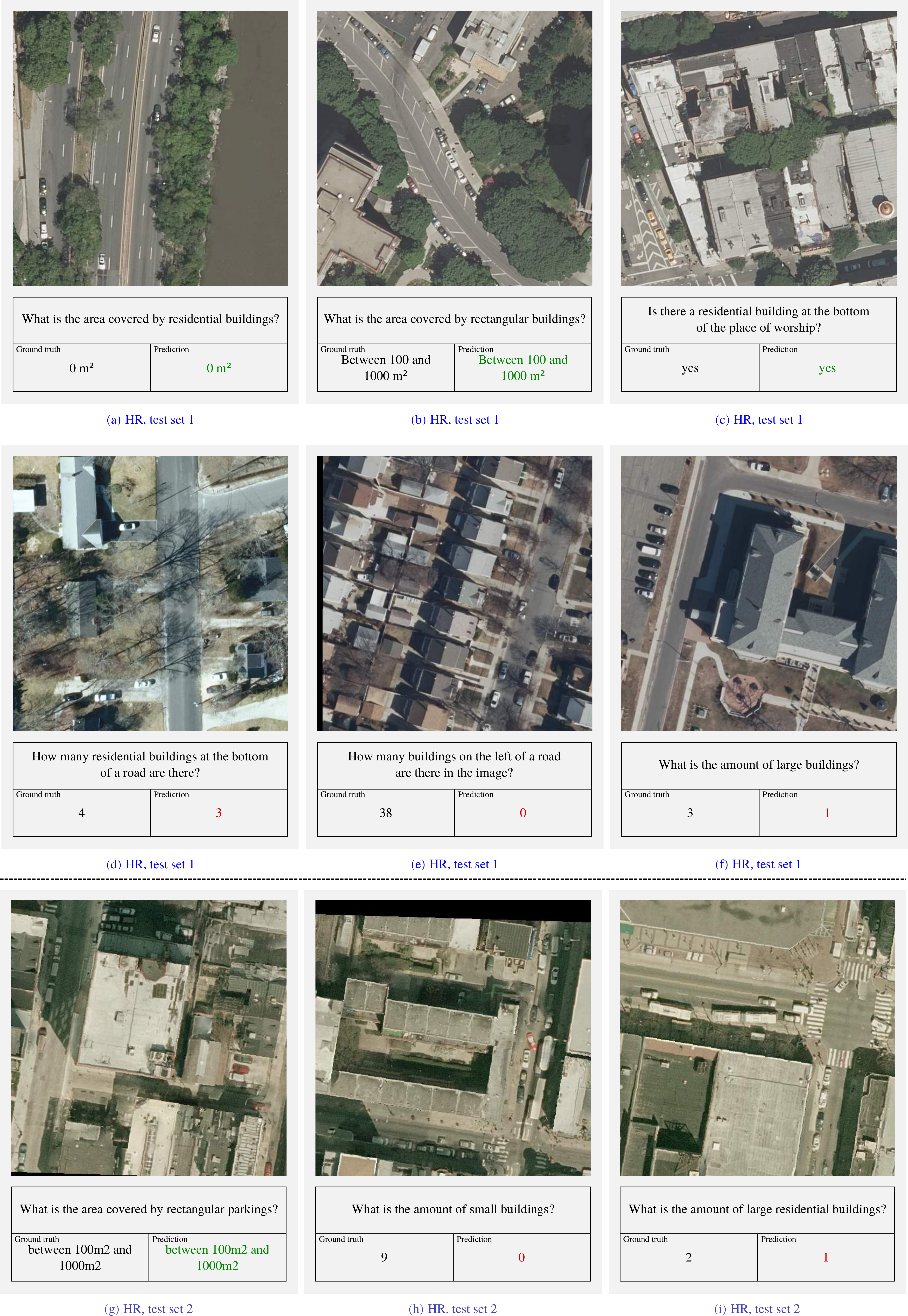}
    \caption{\label{fig:test_visualHR}Samples from the high resolution test sets: (a)-(f) are from the first set of the HR dataset, (g)-(i) are from the second set of the HR dataset.}
\end{figure*}

\begin{figure}
    \centering
    \includegraphics[width=\columnwidth]{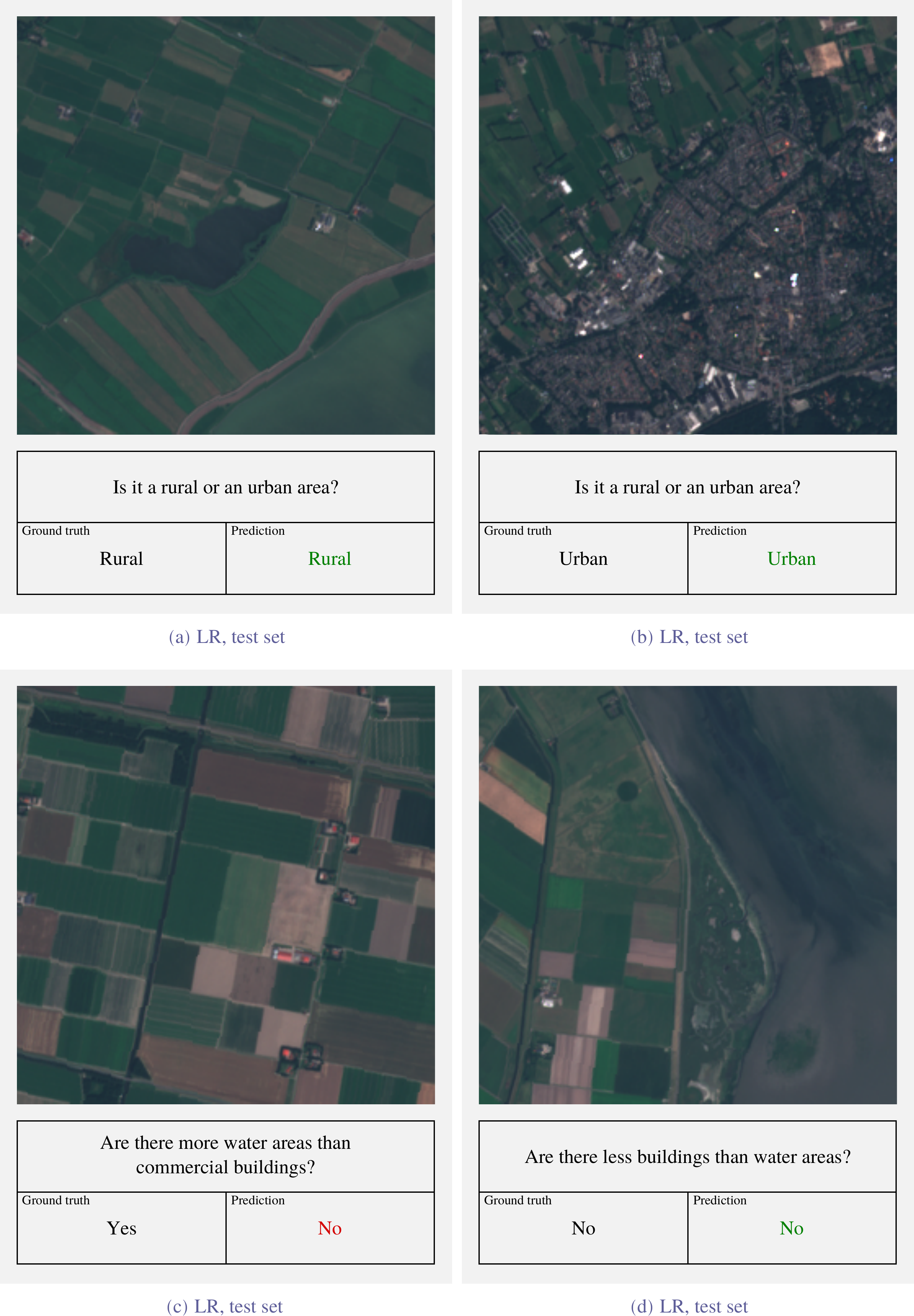}
    \caption{\label{fig:test_visualLR}Samples from the low resolution test set.}
\end{figure}

\begin{figure}
    \centering
    \includegraphics[width=.8\columnwidth]{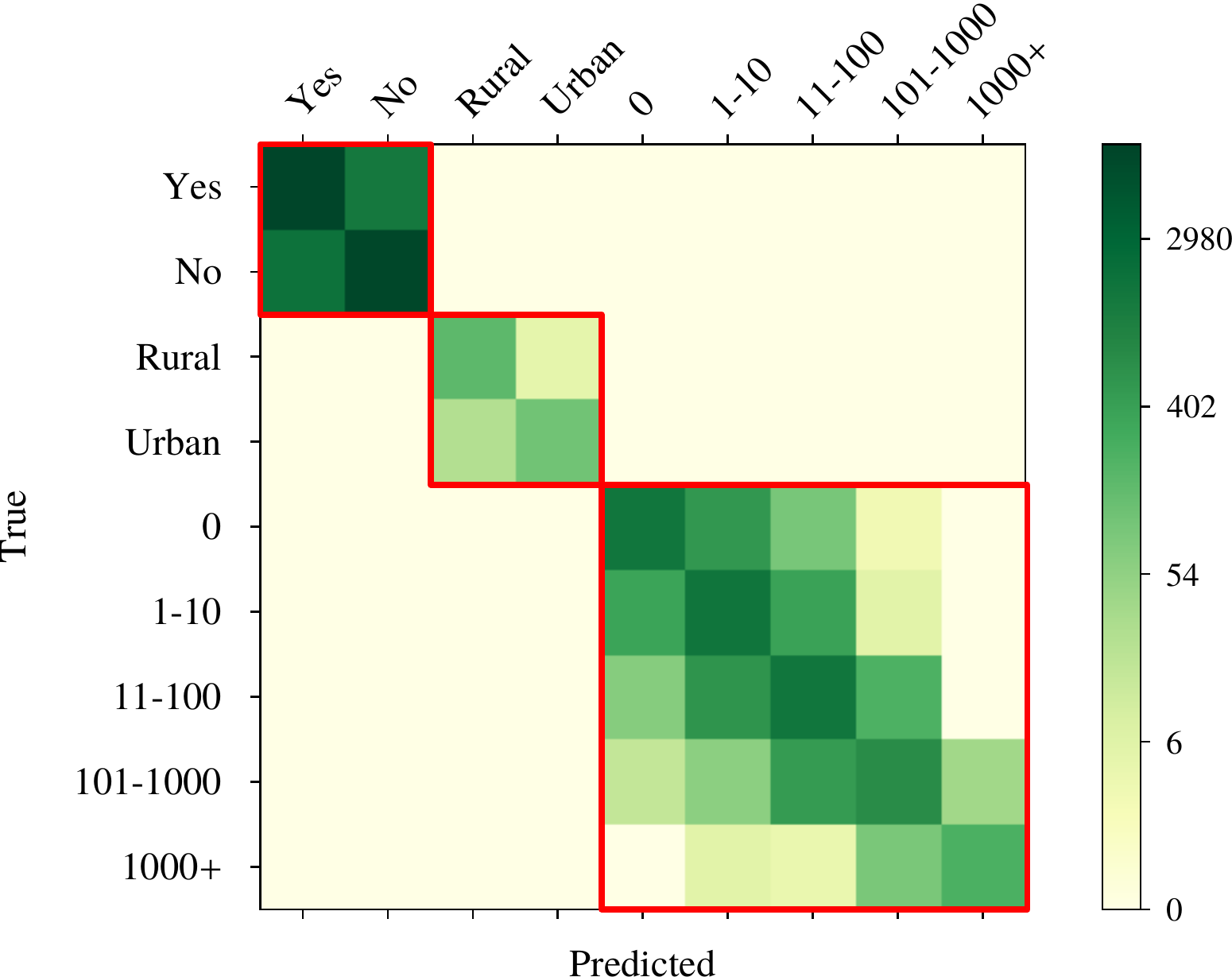}
    \caption{\label{fig:confusionmat_LR}Confusion matrix for the low resolution dataset (logarithm scale) on the test set. Red lines group answers by type ("Yes/No", "Rural/Urban", numbers).}
\end{figure}

\begin{figure*}
    \centering
    \subfloat[Test set 1]{\includegraphics[width=.9\columnwidth]{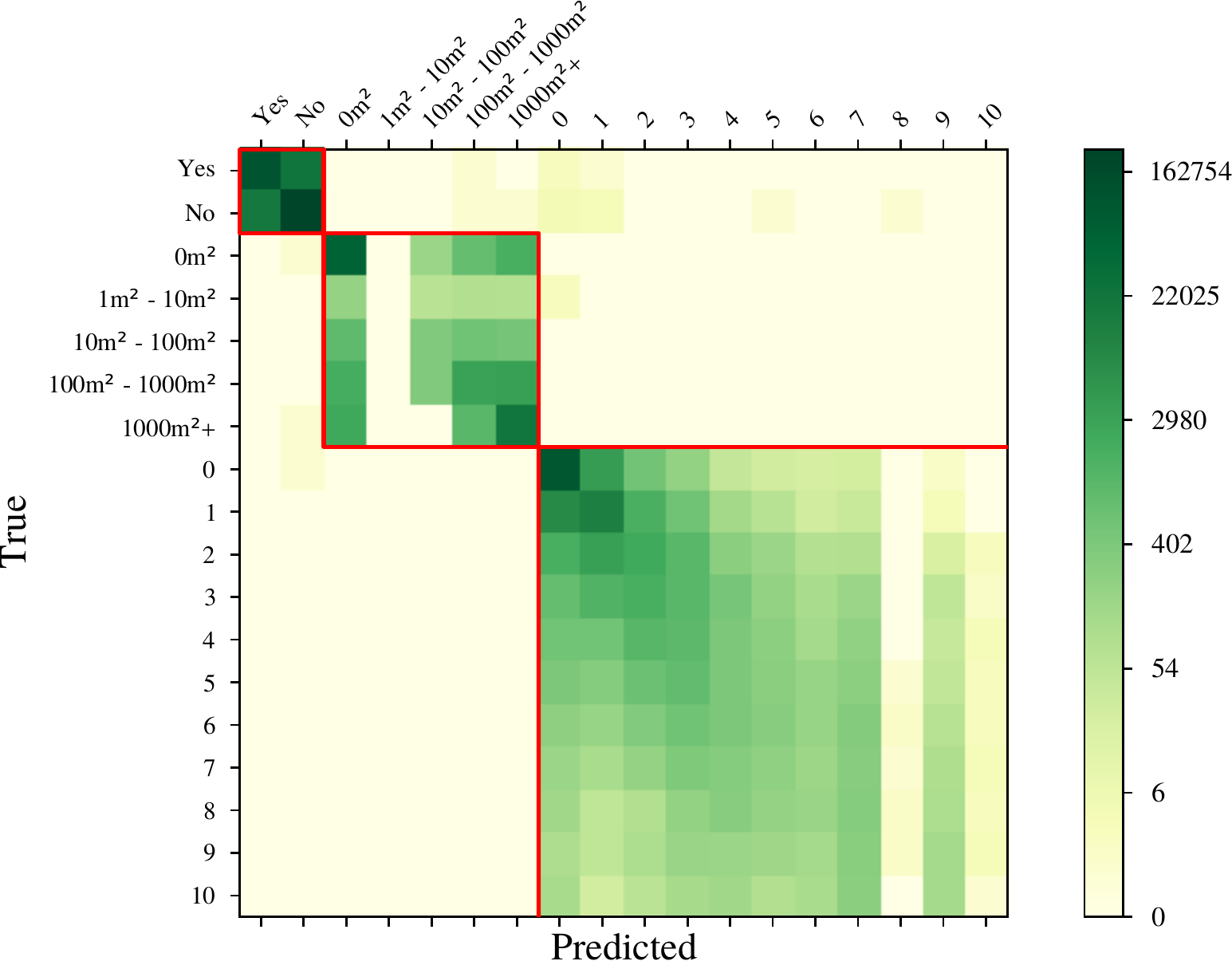}}\hspace{1em}
    \subfloat[Test set 2]{\includegraphics[width=.9\columnwidth]{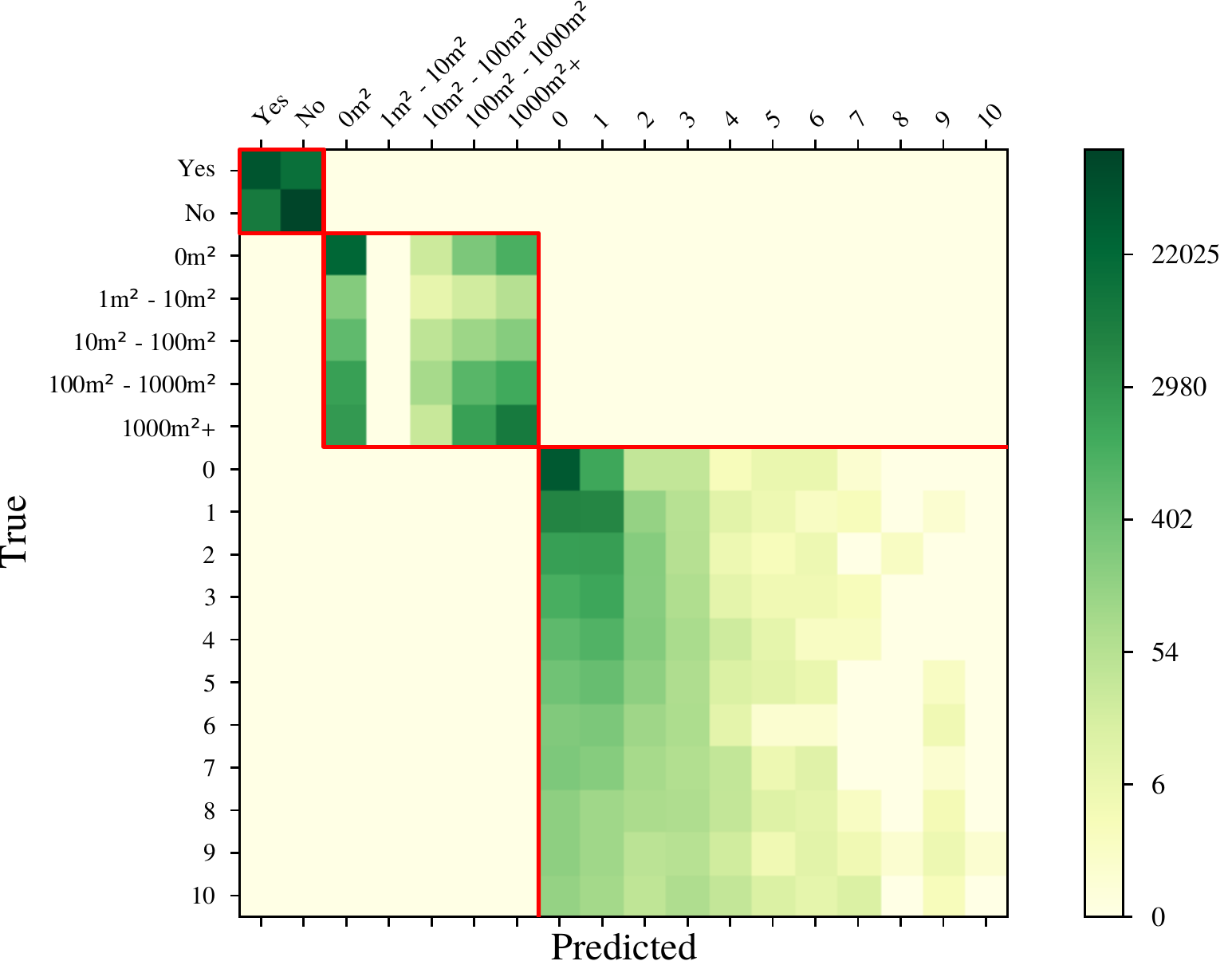}}
    \caption{\label{fig:confusionmat_HR}Subsets of the confusion matrices for the high resolution dataset (counts are at logarithm scale) on both test sets. Red lines group answers by type ("Yes/No", areas, numbers).}
\end{figure*}

We show some predictions of the model on the different test sets in \autoref{fig:test_visualHR} and \autoref{fig:test_visualLR} to qualitatively assess the results. Numerical performance of the proposed model on the LR dataset is reported in \autoref{tab:resultsLR} and the confusion matrix is shown in \autoref{fig:confusionmat_LR}. The performance on both tests sets of the HR dataset are reported in \autoref{tab:resultsHR} and the confusion matrices are shown in \autoref{fig:confusionmat_HR}.\\

\begin{table}[ht]
    \centering
    \caption{\label{tab:resultsLR}Results on the test set of the low resolution dataset. The standard deviation is reported in brackets.}
    \begin{tabular}{|c|c|}
    \hline
    Type        &  Accuracy\\
    \hline
    Count       &  67.01\% (0.59\%)\\
    Presence    &  87.46\% (0.06\%)\\
    Comparison  &  81.50\% (0.03\%)\\
    Rural/Urban &  90.00\% (1.41\%)\\
    \hline
    AA          &  81.49\% (0.49\%)\\
    \hline
    OA          &  79.08\% (0.20\%)\\
    \hline
    \end{tabular}
\end{table}

\begin{table}[ht]
    \centering
    \caption{\label{tab:resultsHR}Results on both test sets of the high resolution dataset. The standard deviation is reported in brackets.}
    \begin{tabular}{|c|c|c|}
    \hline
    Type        &  Accuracy          &  Accuracy\\
                &  Test set 1        &  Test set 2\\
    \hline
    Count       &  68.63\% (0.11\%)  &  61.47\% (0.08\%)\\
    Presence    &  90.43\% (0.04\%)  &  86.26\% (0.47\%)\\
    Comparison  &  88.19\% (0.08\%)  &  85.94\% (0.12\%)\\
    Area        &  85.24\% (0.05\%)  &  76.33\% (0.50\%)\\
    \hline
    AA          &  83.12\% (0.03\%)  &  77.50\% (0.29\%)\\
    \hline
    OA          &  83.23\% (0.02\%)  &  78.23\% (0.25\%)\\
    \hline
    \end{tabular}
\end{table}

\vspace{1em}
\noindent\textbf{General accuracy assessment:}\\
The proposed model achieves an overall accuracy of 79\% on the low resolution dataset (see \autoref{tab:resultsLR}) and of 83\% on the first test set of the high resolution dataset (\autoref{tab:resultsHR}), indicating that the task of automatically answering question based on remote sensing images is possible. When looking at the accuracies per question type (in Tables~\ref{tab:resultsLR} and \ref{tab:resultsHR}), it can be noted that the model performs inconsistently with respect to the task the question is tackling: while a question about the presence of an object is generally well answered (87.46\% in the LR dataset, 90.43\% in the first test set of the HR dataset), counting questions gives poorer performances (67.01\% and 68.63\% respectively). This can be explained by the fact that presence questions can be seen as simplified counting questions to which the answers are restricted to two options: "0" or "1 or more". Classical VQA models are known to struggle with the counting task \cite{zhang2018learning}. An issue which partly explains these performances in the counting task is the separation of connected instances. This problem has been raised for the case of buildings in \cite{JURSECount} and is illustrated in \autoref{fig:test_visualHR}(f), where the ground truth is indicating three buildings, which could also be only one. We found another illustration of this phenomenon in the second test set in \autoref{fig:test_visualHR}(i). This issue mostly arises when counting roads or buildings.

Thanks to the answers' quantization, questions regarding the areas of objects are generally well answered with an accuracy of 85.24\% in the first test set of the HR dataset. This is illustrated in Figures \ref{fig:test_visualHR}(a,b), where presence of buildings (by the mean of the covered area) is well detected. 

However, we found that our model performs poorly with questions regarding the relative positions of objects, such as those illustrated in Figures \ref{fig:test_visualHR}(c-e). While \autoref{fig:test_visualHR}(c) is correct, despite the question being difficult, \autoref{fig:test_visualHR}(d) shows a small mistake from the model and \autoref{fig:test_visualHR}(e) is completely incorrect. These problems can be explained by the fact that the questions are on high semantic level and therefore difficult for a model considering a simple fusion scheme, as the one presented in \autoref{sec:model}.\\

Regarding the low resolution dataset, rural/urban questions are generally well answered (90\% of accuracy), as shown in \autoref{fig:test_visualLR}(a,b). Note that the ground truth for this type of questions is defined as a hard threshold on the number of buildings, which causes an area as the one shown in \autoref{fig:test_visualLR}(b) to be labeled as urban.

However, the low resolution of Sentinel-2 images can be problematic when answering questions about relatively small objects. For instance, in Figures \ref{fig:test_visualLR}(c,d), we can not see any water area nor determine the type of buildings, which causes the model's answer to be unreliable.\\

\vspace{1em}
\noindent\textbf{Generalization to unseen areas:}\\
The performances on the second test set of the HR dataset show that the generalization to new geographic areas is problematic for the model, with an accuracy drop of approximately 5\%. This new domain has a stronger impact on the most difficult tasks (counting and area computation). This can be explained when looking at Figures \ref{fig:test_visualHR}(g-i). We can see that the domain shift is important on the image space, as a different sensor was used for the acquisition. Furthermore, the urban organization of Philadelphia is different from that of the city of New York. This causes the buildings to go undetected by the model in \autoref{fig:test_visualHR}(h), while the parkings can still be detected in \autoref{fig:test_visualHR}(g) possibly thanks to the cars. This decrease in performance could be reduced by using domain adaptation techniques. Such a method could be developed for the image space only (a review of domain adaptation for remote sensing is done in \cite{7486184}) or at the question/image level (see \cite{chao2018cross}, which presents a method for domain adaptation in the context of VQA).\\

\vspace{1em}
\noindent\textbf{Answer's categories:}\\
The confusion matrices indicate that the models generally provide logical answers, even when making mistakes (e.g. it might answer "yes" instead of "no" to a question about the presence of an object, but not a number). Rare exceptions to this are observed for the first test set of the HR dataset (see \autoref{fig:confusionmat_HR}(a)), on which the model gives 23 illogical answers (out of the 316941 questions of this test set).\\

\vspace{1em}
\noindent\textbf{Language biases:}\\
A common issue in VQA models, raised in \cite{balanced_vqa_v2}, is the fact that strong language biases are captured by the model. When this is the case, the answer provided by the model mostly depends on the question, rather than on the image. To assess this, we evaluated the proposed models by randomly selecting an image from the test set for each question. We obtained an overall accuracy of 73.78\% on the LR test set, 73.78\% on the first test set of the HR dataset and 72.51\% on the second test set. This small drop of accuracy indicates that indeed, the models rely more on the questions than on the image to provide an answer. Furthermore, the strongest drop of accuracy is seen on the HR dataset, indicating that the proposed model extracts more information from the high resolution data.

\begin{figure}
    \centering
    \includegraphics[width=.7\columnwidth]{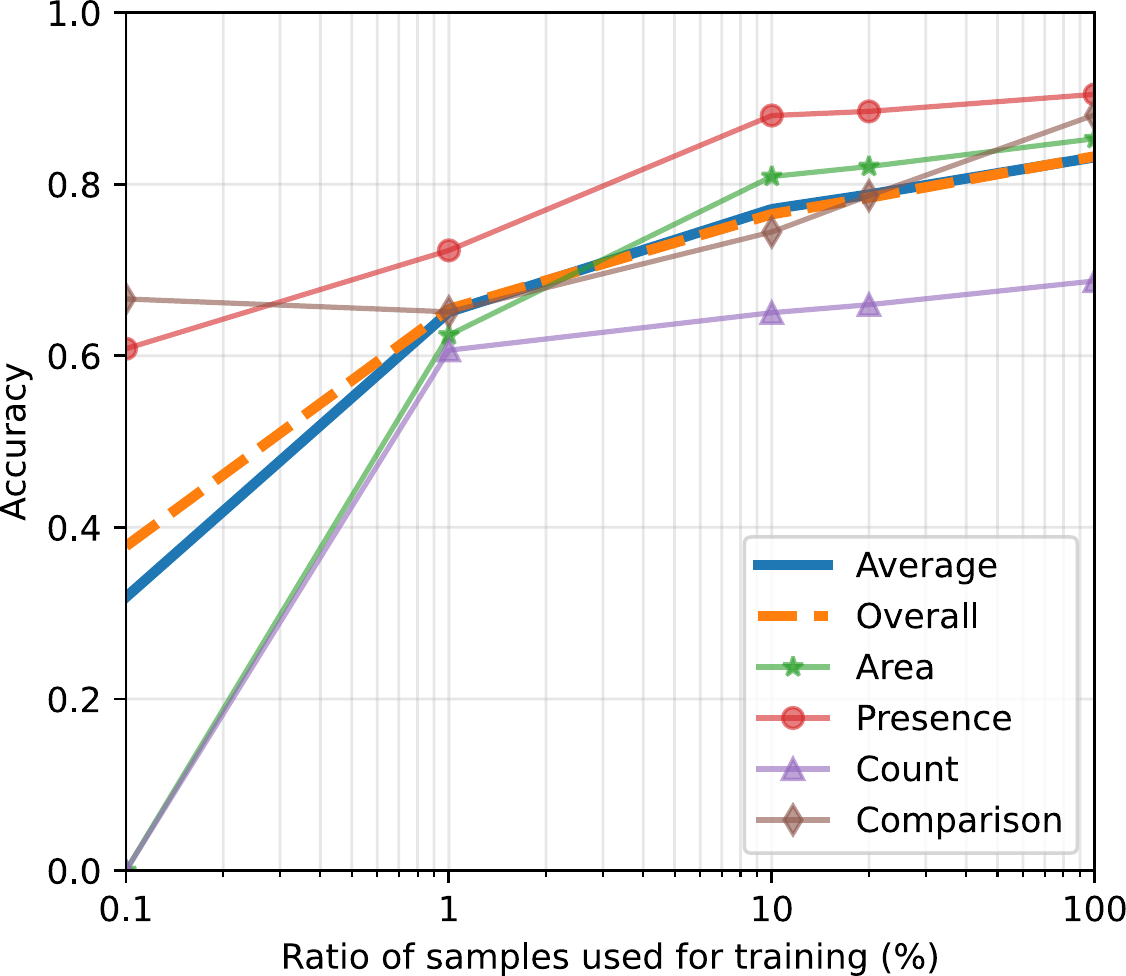}
    \caption{\label{fig:evolVQA}Evolution of the accuracies (evaluated on the first HR test set) after training with subsets of different size of the HR training set.}
\end{figure}
\vspace{1em}
\noindent\textbf{Importance of the number of training samples:}\\
We show in \autoref{fig:evolVQA} the evolution of the accuracies when the model is trained with a fraction of the HR training samples. When using only 1\% of the available training samples, the model already gets 65\% in average accuracy (vs 83\% for the model trained on the whole training set). However, it can be seen that, for numerical tasks (counts and area estimation), larger amounts of samples are needed to achieve the performances reported in \autoref{tab:resultsHR}. This experiment also shows that the performances start to plateau after 10\% of the training data is used: this indicates that the proposed model would not profit substantially from a larger dataset.

\vspace{1em}
\noindent\textbf{Restricted set of questions:}\\
While not appearing in the numerical evaluation, an important issue with our results is the relative lack of diversity in the dataset. Indeed, due to the source of our data (OSM), the questions are only on a specific set of static objects (e.g. buildings, roads, \ldots). Other objects of interest for applications of a VQA system to remote sensing would also include different static objects (e.g. thatched roofs mentioned in \autoref{sec:intro}), moving objects (e.g. cars), or seasonal aspects (e.g. for crop monitoring). Including these objects would require another source of data, or manual construction of question/answer pairs.

Another limitation comes from the dataset construction method described in \autoref{ssec:dataset_method}. We defined five types of questions (count, comparison, presence, area, rural/urban classification). However, they only start to cover the range of questions which would be of interest. For instance, questions about the distance between two points (defined by textual descriptions), segmentation questions (e.g. "where are the buildings in this image?") or higher semantic level question (e.g. "does this area feel safe?") could be added.\\

While the first limitation (due to the data source) could be tackled using other databases (e.g. from national institutes) and the second limitation (due to the proposed method) could be solved by adding other question construction functions to the model, it would be beneficial to use human annotators using a procedure similar to \cite{VQA} to diversify the samples.

\section{Conclusion}
We introduce the task of Visual Question Answering from remote sensing images as a generic and accessible way of extracting information from remotely sensed data. We present a method for building  datasets for VQA, which can be extended and adapted to different data sources, and we proposed two datasets targeting different applications. The first dataset uses Sentinel-2 images, while the second dataset uses very high resolution (30cm) aerial orthophotos from USGS.

We analyze these datasets using a model based on deep learning, using both convolutional and recurrent neural networks to analyze the images and associated questions. The most probable answer from a predefined set is then selected.\\

This first analysis shows promising results, suggesting the potential for future applications of such systems. These results outline future research directions which are needed to overcome language biases and difficult tasks such as counting. The former can be tackled using an attention mechanism \cite{DBLP:journals/corr/WuTWSDH16}, while the latter could be tackled by using dedicated components for counting questions \cite{JURSECount} in a modular approach.

Issues regarding the current database raised in \autoref{sec:results} also need to be addressed to obtain a system capable of answering a more realistic range of questions. This can be done by making the proposed dataset construction method more complex or by using human annotators.

\section*{Acknowledgment}
The authors would like to thank CNES for the funding of this study (R\&T project "Application des techniques de Visual Question Answering \`a des données d'imagerie satellitaire").

{\small 
\bibliographystyle{IEEEtran}
\bibliography{Master}
}

\begin{IEEEbiography}[{\includegraphics[width=1in,height=1.25in,clip,keepaspectratio]{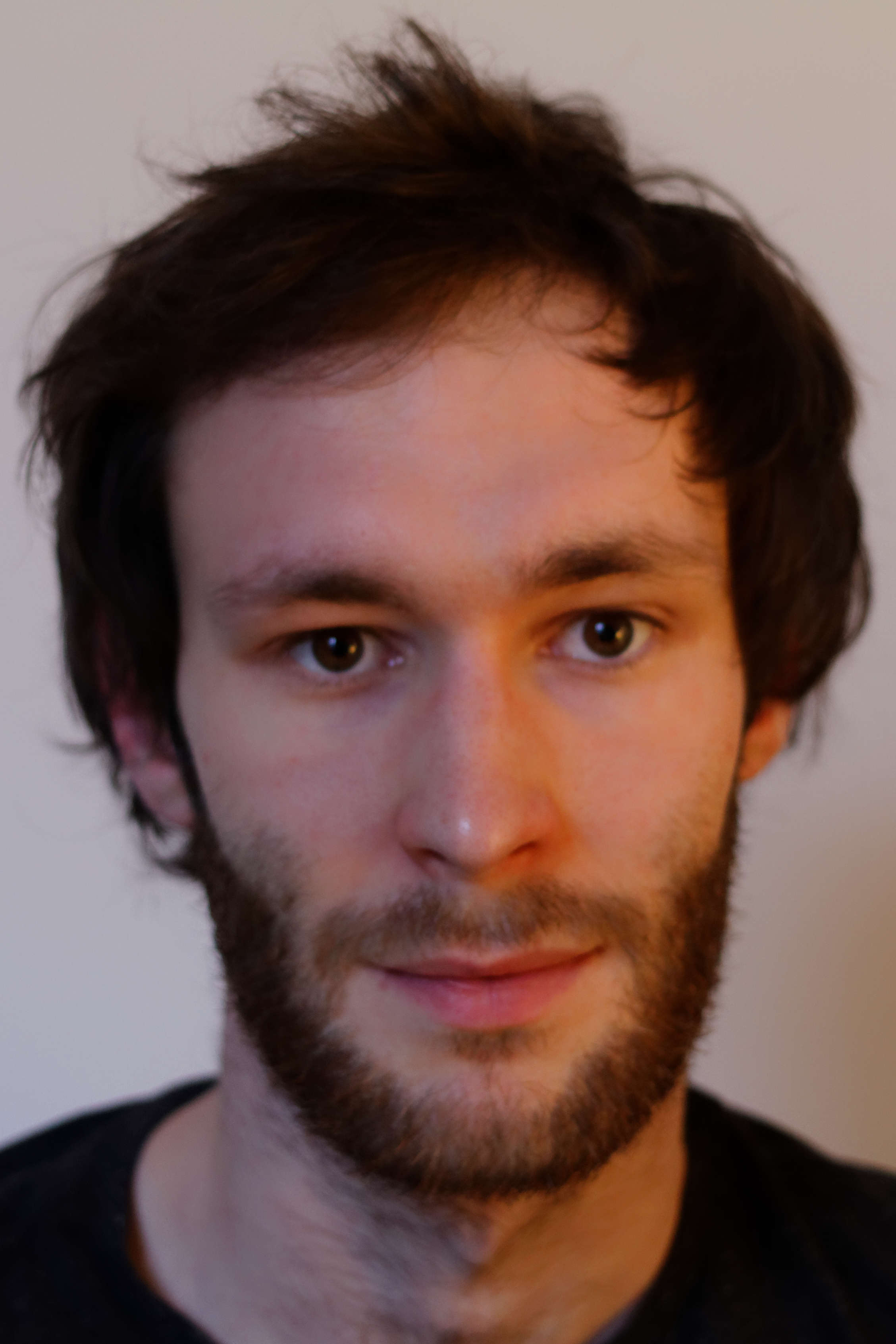}}]{Sylvain Lobry}
(S’16 - M’17) received the Engineering degree  from Ecole pour l'Informatique et les Techniques Avanc\'ees (EPITA), Kremlin Bic\^etre, France in 2013, the Master’s degree in science and technology  from  the  University  Pierre et Marie Currie (Paris 6), Paris in 2014, and the Ph.D. degree from Telecom Paris, Paris, France, in 2017. He is currently a Post-Doctoral Researcher with the Geo-Information Science and Remote Sensing Laboratory, Wageningen University, The Netherlands. His research interests include radar image processing and multimodal processing of heterogeneous satellite data.
\end{IEEEbiography}
\begin{IEEEbiography}[{\includegraphics[width=1in,height=1.25in,clip,keepaspectratio]{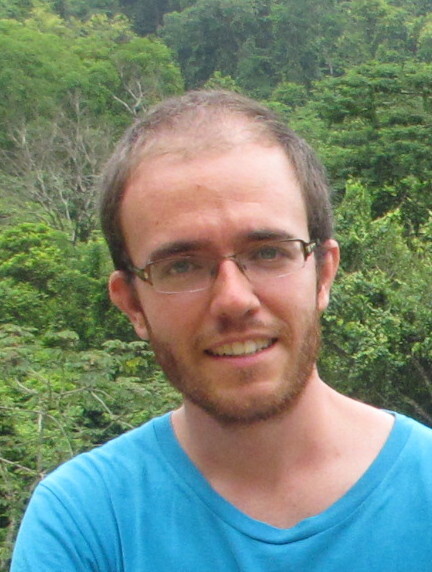}}]{Diego Marcos}
Diego Marcos obtained an MSc degree on Computational Sciences and Engineering from the Ecole Polytechnique F\'ed\'erale de Lausanne, Switzerland, in 2014 and a Ph.D. degree in environmental sciences from Wageningen University, The Netherlands, in 2019. He is currently a Post-Doctoral Researcher with the Geo-Information Science and Remote Sensing Laboratory, at Wageningen University. His research interests include computer vision and deep learning interpretability applied to geospatial data.
\end{IEEEbiography}
\begin{IEEEbiography}[{\includegraphics[width=1in,height=1.25in,clip,keepaspectratio]{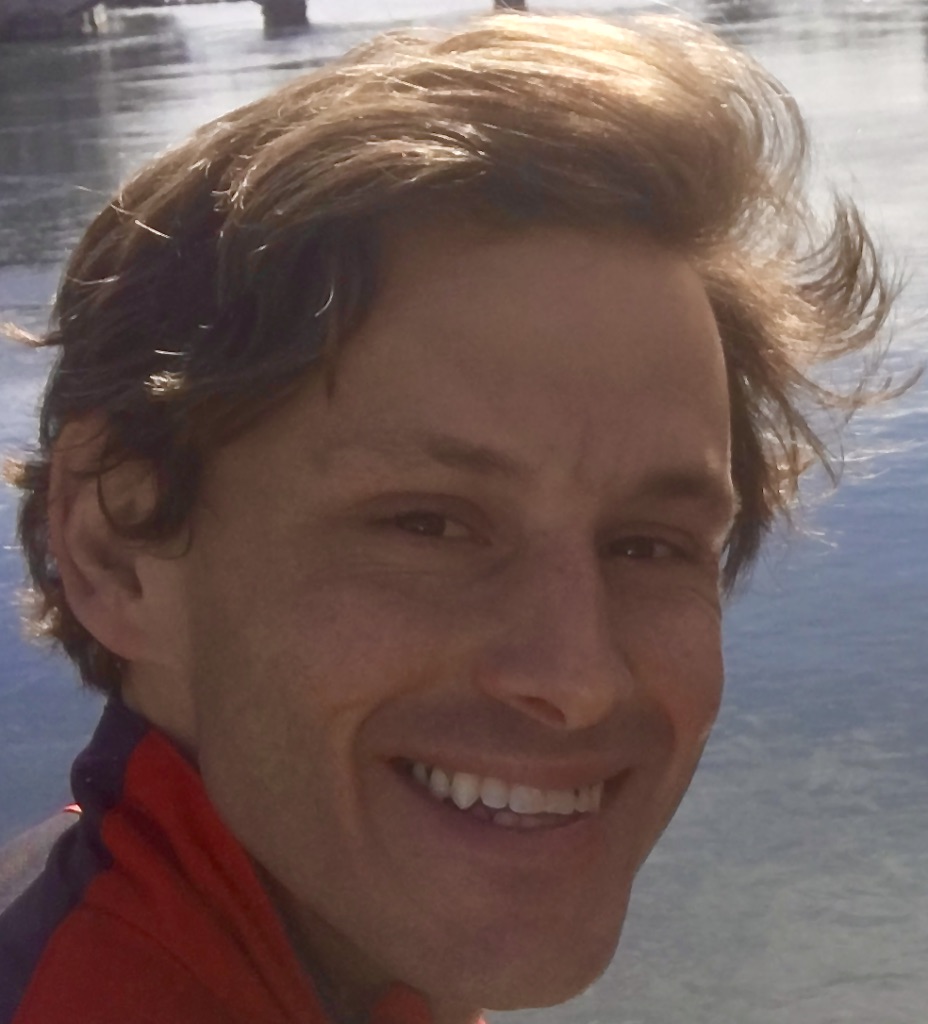}}]{Jesse Murray}
received an MSc degree in Geo-Information Science from Wageningen University, Wageningen, The Netherlands, in 2019. He is currently a Ph.D. candidate with the Geodetic Engineering Laboratory, at the Ecole Polytechnique Fédérale de Lausanne, Switzerland. His research interests include image processing and 3D geometry reconstruction using computer vision and spatio-temporal data.
\end{IEEEbiography}
\begin{IEEEbiography}[{\includegraphics[width=1in,height=1.25in,clip,keepaspectratio]{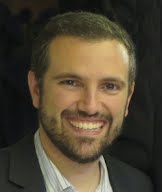}}]{Devis Tuia}
Devis Tuia (S’07 - M’09 - SM’15) received the Ph.D. degree from the University of Lausanne, Lausanne, Switzerland, in 2009. He was a Post-Doctoral Researcher in Valéncia, Boulder, CO, USA, and \'Ecole polytechnique f\'ed\'erale de Lausanne (EPFL), Lausanne. From 2014 to 2017, he was an Assistant Professor with the University of Zurich, Zürich, Switzerland. He is currently a Full Professor with the Geo-Information Science and Remote Sensing Laboratory, Wageningen University, Wageningen, The Netherlands. His research interests include algorithms for data fusion of geospatial data (including remote sensing) using machine learning and computer vision.
\end{IEEEbiography}

\end{document}